\pdfoutput=1
\documentclass[11pt]{article}

\usepackage[preprint]{acl}

\usepackage{times}
\usepackage{latexsym}

\usepackage[T1]{fontenc}
\usepackage[utf8]{inputenc}

\usepackage{microtype}
\usepackage{graphicx}
\usepackage{array}
\usepackage{booktabs}
\usepackage{multirow}
\usepackage{multicol}
\usepackage{color}
\usepackage{threeparttable}
\usepackage{amsmath}
\usepackage[utf8]{inputenc}
\usepackage{longtable}
\usepackage{amssymb}
\usepackage{makecell}
\usepackage{xspace}
\usepackage{enumitem}
\usepackage{amsthm}
\usepackage{amssymb, amsfonts}
\usepackage{algorithm}
\usepackage{algpseudocode}
\usepackage{mathrsfs}
\definecolor{Gray}{gray}{0.95}
\definecolor{hookersgreen}{rgb}{0.0, 0.44, 0.0}
\definecolor{indiagreen}{rgb}{0.07, 0.53, 0.03}
\definecolor{islamicgreen}{rgb}{0.0, 0.56, 0.0}
\definecolor{kellygreen}{rgb}{0.3, 0.73, 0.09}
\definecolor{alizarin}{rgb}{0.82, 0.1, 0.26}
\usepackage{url}
\definecolor{mydarkblue}{rgb}{0,0.08,0.45}
\definecolor{mycitecolor}{HTML}{2980b9}
\definecolor{mylinkcolor}{HTML}{c0392b}
\definecolor{citecolor}{HTML}{0071BC}
\usepackage{hyperref}
\usepackage[normalem]{ulem}
\usepackage{pifont}% http://ctan.org/pkg/pifont
\newcommand{\cmark}{{\color{kellygreen} \ding{51}}}
\newcommand{\xmark}{{\color{alizarin} \ding{55}}}

\usepackage{inconsolata}
\usepackage{lipsum}
\usepackage[noabbrev,capitalise]{cleveref}

\definecolor{mydarkgreen}{rgb}{0.2,0.7,0.2}
\def \ourmethod {ETR\xspace}

\title{An Expert is Worth One Token: Synergizing Multiple Expert LLMs as Generalist via Expert Token Routing}

\newcommand{\zju}{1}%
\newcommand{\hr}{2}%
\newcommand{\bd}{3}%

\author{%
Ziwei Chai\textsuperscript{\zju,\hr},
Guoyin Wang\textsuperscript{\bd},
Jing Su\textsuperscript{\bd},
Tianjie Zhang\textsuperscript{\zju},
Xuanwen Huang\textsuperscript{\zju},
Xuwu Wang\textsuperscript{\bd} \\
\textbf{Jingjing Xu}\textsuperscript{\bd},
\textbf{Jianbo Yuan}\textsuperscript{\bd},
\textbf{Hongxia Yang}\textsuperscript{\bd},
\textbf{Fei Wu}\textsuperscript{\zju},
\textbf{Yang Yang}\textsuperscript{\zju}\thanks{~~Corresponding authors.} \\
\textsuperscript{1} Zhejiang University, \textsuperscript{2} Beijing Huairou Laboratory, \textsuperscript{3} ByteDance Inc.\\
\texttt{
\{zwchai, yangya\}@zju.edu.cn 
}
}

\begin{document}
\maketitle
\begin{abstract}
We present Expert-Token-Routing, a unified generalist framework that facilitates seamless integration of multiple expert LLMs. Our framework represents expert LLMs as special expert tokens within the vocabulary of a meta LLM. The meta LLM can route to an expert LLM like generating new tokens. Expert-Token-Routing not only supports learning the implicit expertise of expert LLMs from existing instruction dataset but also allows for dynamic extension of new expert LLMs in a plug-and-play manner. It also conceals the detailed collaboration process from the user's perspective, facilitating interaction as though it were a singular LLM. Our framework outperforms various existing multi-LLM collaboration paradigms across benchmarks that incorporate six diverse expert domains, demonstrating effectiveness and robustness in building generalist LLM system via synergizing multiple expert LLMs. \footnote{Codes are available at \href{https://github.com/zjunet/ETR}{https://github.com/zjunet/ETR}.}
\end{abstract}

\section{Introduction}

Large language models (LLMs)—notably, GPT-4 and LLaMA~\citep{touvron2023llama}—have demonstrated remarkable capabilities across a wide spectrum of tasks. However, their performance and reliability in certain specialized domains sometimes still fall short of expectations~\citep{wang2023survey,zhao2024loraretriever}. Motivated by this, the recent year has seen emergence in the development of "expert" LLMs, with a particular focus on tailoring a general LLM to excel in specific domains or tasks. For example, there are efforts to refine the pretraining corpus to develop expert LLMs with enhanced capabilities in specific domains, such as CodeLlama~\citep{rozière2024code} and DeepSeekMath~\citep{shao2024deepseekmath}. Furthermore, applying continual training or supervised fine-tuning (SFT) on specially selected corpora can lead to significant enhancements on targeted tasks like data analysis~\citep{li2023llm,hu2024infiagentdabench} and financial analysis~\citep{chen2023disc}, making a general LLM into a specialized expert.

In this study, we focus on the following specific research question: \textit{how can we synergize various expert LLMs into a singular generalist framework?} The practicality of this question lies in the fact that it's unrealistic to ask users to precisely switch between various expert LLMs, given the diversity and subtle difference in their capabilities. There are mainly two paradigm to build such a generalist LLM system, termed in this paper as the prompt-based methods and the router-based methods. Prompt-based methods~\citep{li2023camel,wang2024unleashing,suzgun2024metaprompting} treat expert LLMs as conversational agents. A "meta" LLM guided by crafted instruction communicates with expert LLMs to acquire domain-specific information. For the router-based methods~\citep{jiang2023llmblender,lu2023routing}, a router model is trained to route each query to decide which expert is responsible to answer it.

However, both of the aforementioned paradigms have some notable limitations (\cref{tab:motivation}). 
When employing the prompt-based method, successful expert collaboration significantly relies on how the instruction are crafted. For cases where the expertise of expert LLMs is implicit or  undisclosed, pinpointing relevant expert knowledge through prompts could be inefficient. In the other hand, router-based methods requires training an extra router model tailored to the current expert LLM library, which needs to be retrained whenever a new expert LLM is added.

In this work, we introduce a novel technique termed Expert-Token-Routing (\ourmethod). \ourmethod employs a multi-expert LLM collaboration framework, wherein a meta LLM can seamlessly switch to a specific expert LLM. The core idea of \ourmethod is encoding expert LLMs as special tokens within the vocabulary of the meta LLM. Inspired by ToolkenGPT~\citep{hao2023toolkengpt}, which captures tool semantics using special tokens, \ourmethod employs expert tokens to characterize the expertise of different expert LLMs. Under this formulation, the meta LLM can route to an expert LLM like generating new tokens. This approach allows for expert routing without training extra router model, simply by appending trained expert tokens into a frozen meta LLM. To train the expert tokens, we construct an automated pipeline to collect expert queries from existing datasets. \ourmethod can learn expert token embeddings from collected expert queries, which profiles the strengths of expert LLMs, thereby handling situations where the expertise of expert LLMs is implicit or undisclosed.

Our framework also supports a plug-and-play integration for new expert LLMs, allowing providers to add their expert LLMs and trained expert tokens to the framework as modular plug-ins, eliminating the need for modifications to any other part of the framework. Another appealing characteristic of our method is that it conceals the collaboration process between the meta LLM and expert LLMs from the user's perspective, facilitating interaction with the framework as though it were a singular LLM.

In our comprehensive experiments across six different domains, we compare \ourmethod with the existing paradigms for expert LLM collaboration. Our findings indicate that ETR significantly outperforms the baselines as a generalist, with an overall performance improvement of 5.64\%. It is worth noting that our framework also integrates the advantages of existing paradigms, as illustrated in \cref{tab:motivation}.

The core contribution of this work is the introduction of a singular generalist framework that
facilitates seamless integration and collaboration of expert LLMs. This framework not only supports learning the implicit expertise of expert LLMs from existing instruction dataset but also allows for dynamic extension of new expert LLMs in a plug-and-play manner. The effectiveness of our framework is showcased across benchmarks that incorporate six diverse expert domains.

\begin{table}[t]
\caption{
Comparison of different expert LLM collaboration paradigms. Implicit Expertise means it's difficult to obtain a precise depiction of the expertise of an expert LLM. Dynamic Extension is about whether an expert LLM can be incorporated in a plug-and-play manner without making changes to the rest of the framework.}
\centering
\small
\resizebox{\linewidth}{!}{ \renewcommand{\arraystretch}{1.0}
\begin{tabular}{@{}l c c c c@{}}
\toprule 
\makecell{Expert Collaboration\\ Paradigm} & \makecell{Extra Model \\ Free} & \makecell{{Implicit}\\{Expertise}} &  \makecell{{Dynamic}\\{Extension}}  & \makecell{Transparent \\ to Users}  \\ \midrule
Prompt-based & \cmark & \xmark & \cmark & \xmark  \\
Router-based & \xmark & \cmark & \xmark & \cmark \\  \midrule
\ourmethod ({\bf Ours}) & \cmark  & \cmark &   \cmark  &  \cmark \\
\bottomrule
\end{tabular}}
\label{tab:motivation}
\end{table}
\section{Approach}

In this section, we present \ourmethod, which enables the integration of multiple expert LLMs into a singular generalist LLM framework for leveraging expert knowledge without the need for heavily fine-tuning or designing crafted instruction templates. Typically, an expert LLM $\epsilon$ is derived from base pre-trained LLMs through continual pretraining or supervised fine-tuning (SFT) on data of target domain. By careful data selection, an expert LLM of smaller scale can exceed a larger, general-purpose LLM in specific areas like coding~\citep{guo2024deepseekcoder} or math problem-solving~\citep{shao2024deepseekmath}. 

Given an expert LLM library $E = \{ \epsilon_1, \epsilon_2, \ldots \}$, our goal is to develop an LLM framework which can utilizing the expert knowledge available in the library $E$ when solving problems within specific domains. To determine when and which expert to corporate with, we use a general-purpose LLM $\Omega$ as a meta LLM. And expert LLM $\epsilon_i$ is represented by a special token (expert token) witin the meta LLM's vocabulary.

\subsection{Framework Overview}

\begin{figure*}[htbp]
    \centering
    \includegraphics[width=\linewidth]{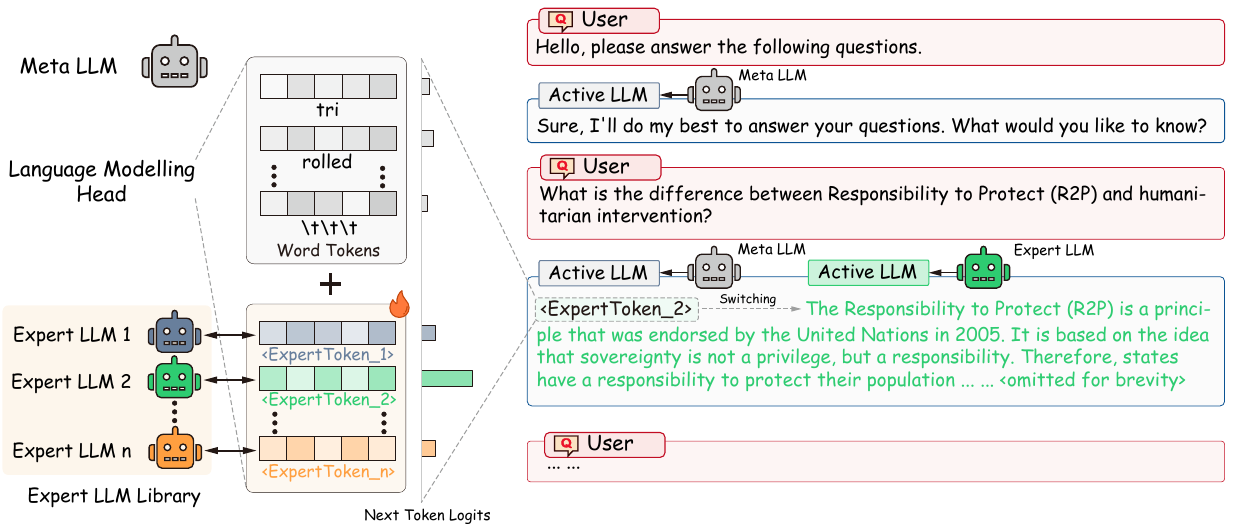}
    \caption{The framework of \ourmethod. During the \ourmethod's decoding process, the meta LLM serves as the default active LLM. When predicting the expert token, the active LLM switches to the corresponding expert LLM. The Expert tokens are appended to the frozen language modeling head of the meta LLM, where it is treated equally with word tokens during the next token prediction.}
    \label{fig:model}
\end{figure*}

The core design of \ourmethod is to encode expert LLMs as special tokens (termed "expert tokens") in the meta LLM's vocabulary. In this formulation, the expert routing is redefined within the LLM's standard next word prediction paradigm. An expert LLM is activated for decoding once its expert token is predicted as the next token by the meta LLM. The expert tokens are parameterized as an embedding matrix $W_E \in \mathbb{R}^{|E|\times d}$ and are appended to the language model head $W_V$ of the meta LLM like regular word tokens.

Assuming we've already trained expert token embeddings $W_E$ (to be described in \cref{subsec:learning}), we will start by introducing how our framework works in inference. As shown in \cref{fig:model}, the meta LLM is served as the gateway for handling queries by default. And it is designed to treat word tokens and expert tokens uniformly. Specifically, the language modeling head of the meta LLM is a concatenation of original word token head and expert embeddings. Consequently, the meta LLM generates the response by predicting the next token with the following probability:
\begin{equation}
P_\Omega(t_i | t_{<i}) = \text{softmax}([W_V; W_E] \cdot \mathbf{h}_{i-1}) 
\end{equation}
where $W_V \in \mathbb{R}^{|V|\times d}$ and $|V|$ is the vocabulary size of word tokens. Note that the next token can be either a word token or a expert token, i.e. $t_i \in V \cup E$.

Once an expert token is predicted, the meta LLM halts decoding. Subsequently, the corresponding expert LLM continues decoding, using the already generated context as its prefix. \cref{alg:algo1} provides a detailed description of the inference procedure of \ourmethod. Our framework operates as a singular LLM, making the underlying expert collaboration invisible to the user. As a result, interacting with our framework is indistinguishable from interacting with a general LLM.

\begin{algorithm}[h]
\caption{Inference procedure of \ourmethod}
\label{alg:algo1}
\begin{algorithmic}[1]
\State \textbf{Input:} Query $q$, expert LLM Library $E = \{\epsilon_1, \epsilon_2, \ldots\}$, meta LLM $\Omega$
\State \textbf{Output:} Response $r$

\Procedure{Generate}{$q, E, \Omega$}
    \State Initialize context $c \gets q$
    \State Initialize active LLM $\lambda \gets \Omega$ \Comment{Start with meta LLM}
    \While{Response not complete}
        \State $t_i \gets$ Predict next token with $\lambda$ given $c$
        \If{$t_i$ is a word token}
            \State Update context $c \gets c \;||\; t_i$
        \ElsIf{$t_i$ is $\epsilon$'s expert token}
            \State $\lambda \gets \epsilon_j$ \Comment{Switch active LLM to the expert LLM}
        \EndIf
    \EndWhile
    \State $r \gets c$ 
    \State \textbf{return} $r$
\EndProcedure
\end{algorithmic}
\end{algorithm}

\subsection{Learning Expert Embeddings via Expert Query Set}
\label{subsec:learning}

% Our framework freezes the original parameters of the LLM and introduces only a minimal training workload with the addition of expert embeddings. Whenever a new expert is introduced to the expert library, the expert embedding can be conveniently expanded.  

Ideally, when a query is falls within the expertise of a specific expert, the corresponding expert token should be generated. Therefore, we hope that the expert embedding implicitly encodes the expert's strengths.

Drawing on the idea that experts outperform non-experts in addressing problems within their expertise area, we utilize an expert query set $\epsilon_i$ to learn an expert LLM's token embedding. Specifically, we consider a query-response pair that has significantly lower loss on the expert LLM $\epsilon_i$ compared to a general LLM to be $\epsilon_i$'s expert query. Given an instruction dataset $D$ consisting of query-response pairs  $\{(q_j, r_j)_{j=0, 1, \ldots}\}$, the loss of LLM $\epsilon$ on $j$-th query-response pair is defined as

\begin{equation}
L_\epsilon(j) = \sum_{x_k \in r_j} -logP_\epsilon(x_k|q_j)
\end{equation}
where $x_k$ is the tokens in the response. The instruction dataset used for collecting expert query set can be either existing instruction-tuning dataset~\citep{alpaca} or synthetic data generated by LLMs

For each expert LLM $\epsilon_i$ in the expert library, we build an expert query set $\mathcal{A}_i$ according to the following criteria: 

\begin{equation}
    \mathcal{A}_i = \{(q_j, r_j) |L_{\Omega}(j) - L_{\epsilon_i}(j) > \tau\}
\end{equation}
where $L_\Omega$ is the loss of the meta LLM. Therefore, $\mathcal{A}_i$ is the collection of queries that the expert LLM $\epsilon_i$ adept at solving, outperforming the general-purpose meta LLM $\Omega$. In the training process, queries in $\mathcal{A}_i$ serves as prefix and a special expert token $<\texttt{ExpertToken\_i}>$ is appended as ground truth for the next token prediction.
Specifically, the training objective of \ourmethod is:
\begin{equation} 
\mathcal{L}(W_E) = \sum_i^{|E|}\sum_{{q_j} \in \mathcal{A}_i} -logP(\texttt{<ExpertToken\_i> }|q_j) 
\end{equation}

This embedding matrix $W_E$ are the only tunable parameters.
Therefore, similar to pioneering work on expanding the language modeling head of LLMs (such as ToolkenGPT~\citep{hao2023toolkengpt}), training ETR does not require gradient propagation through the main part of the LLM parameters, leading to training that is both more stable and efficient than prompt tuning~\citep{lester2021power} or prefix tuning~\citep{li2021prefixtuning}. Thus, the GPU memory usage for expert token embedding tuning is nearly the same to that of LLM inference.
\section{Experiments}

\subsection{Experimental Settings}

\label{sec:exp_setting}

\paragraph{Datasets} 

We build a dataset comprising a mix of questions from six diverse expert domain. Specifically, we employ GPT-4 to rank the 57 subjects in MMLU~\citep{hendrycks2021measuring} across STEM, the humanities, the social sciences, and more, based on their level of specialization. We select the six highest-ranked subjects, which are astronomy, electrical engineering, security studies, prehistory, international law and human sexuality. Their test question sets are merged to form a dataset, named MMLU-Expert, which includes questions from six different specialized fields.

% We build two datasets comprising questions across from various professional domains.  We choose six highly specialized categories from MMLU~\citep{hendrycks2021measuring} to assemble a dataset, named MMLU-Expert. Also, we construct a mixed dataset by sampling questions from datasets widely used in different reasoning tasks, such as MMLU~\citep{hendrycks2021measuring}, MATH~\citep{hendrycks2021measuringmath} and LeetCode~\citep{guo2024deepseekcoder}.

% \paragraph{Meta and Expert LLM Setups}

% The meta LLM in our framework can be any pre-trained LLMs. In this paper, we choose Qwen, a well-pretrained LLM, as the meta LLM. For every expert domain in the MMLU-Expert, we collect several thousand question-and-answer data. We build an expert LLM library of six expert LLMs by fine-tuning a Qwen-7B-Chat model.

\paragraph{Expert LLM Construction via SFT}

Most previous research~\citep{li2023camel, xu2023expertprompting, liu2023dynamic, suzgun2024metaprompting} create expert LLMs by constructing specialized role prompts (i.e. mathematics, laywer). Although creating experts by role-playing is straightforward, it is difficult to effectively inject domain knowledge into an LLM through just several lines of expert description prompt. We believe that constructing expert LLMs through continual training/supervised fine-tuning is more aligned with real-world applications than the role-playing prompting approach. 

Therefore, we collect a domain expert instruction dataset comprising knowledge across 6 different domains in MMLU-Expert. We synthesize 7.5K question-answer pairs for each domain, by harnessing unnatural instructions~\citep{kervadec2023unnatural} to extract domain knowledge from GPT-4. Specifically, a seed question set is initialized using the MMLU validation questions. In each cycle, three questions are randomly selected from the seed question set, and GPT-4 is tasked with generating a question within the same domain. To avoid the risk of data leakage by GPT-4 reproducing MMLU-Expert test questions from its memory, we filter out questions which have a high similarity (measured by BERTScore~\citep{bert-score} > 0.8) with test questions. We once again utilize GPT-4 to get answers for these generated questions. This approach allows extracting domain knowledge from GPT-4 into a question-and-answer dataset.

% See \cref{fig:data_collection} for the detailed dataset construction process.

\begin{figure}[htbp]
  \centering
  \includegraphics[width=\linewidth]{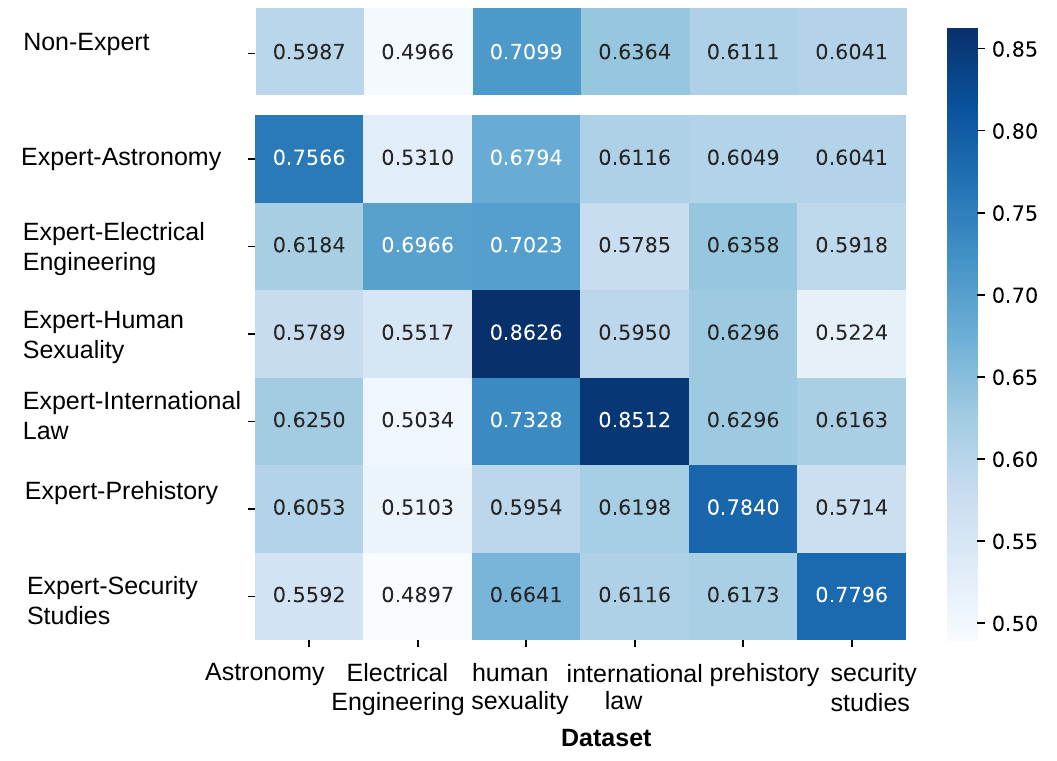}
  \caption{Expert LLMs' accuracy on MMLU-Expert.}
  \label{fig:expert}
\end{figure}

We then apply supervised fine-tuning on the synthetic dataset to get expert LLMs injected with domain knowledge, namely \texttt{Expert-DomainName}. \cref{fig:expert} shows the expert LLMs' performance across different domains of MMLU-Expert. Compared to non-expert models before fine-tuning, the expert models exhibit a significant performance gain of 17.90\% on the in-domain test sets.

%The expertise of each fine-tuned LLMs is evaluated based on their accuracy scores on selected domains of the MMLU test set~\citep{hendryckstest2021}. 

\begin{figure}[htbp]
    \centering
    \includegraphics[width=\linewidth]{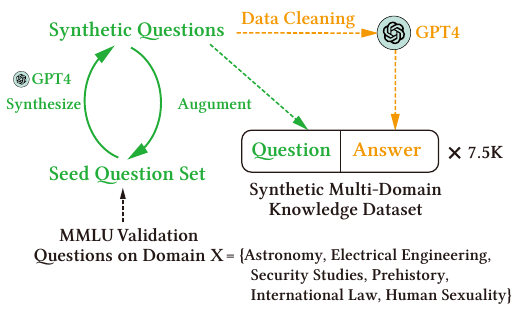}
    \caption{Collection process of multi-domain knowledge dataset synthesized through GPT-4. To prevent potential data leakage, synthetic questions with high BERTScore~\citep{bert-score} to any question in the test set are filtered out.}
    \label{fig:data_collection}
\end{figure}

\paragraph{Evaluation Metric} With access to both an expert LLM library  and a general LLM, the performance of a multi-expert LLM framework are evaluated through the overall accuracy on the MMLU-Expert. For each question, there exists an optimal expert in the expert LLM library. The framework's capability to accurately route questions to these optimal experts serves as a critical measure.

\paragraph{Baselines} We compare our framework with two types of baseline methods: the prompting-based expert collaboration methods and the router-based expert routing methods. It's worthy noting that in the existing prompting-based methods, as exemplified by Meta-Prompting~\citep{suzgun2024metaprompting}, there are two distinct components in its instruction: collaborating with experts and forming expert identity, both of which are crucial to the effectiveness of this method. However, in this paper's setting, an external expert LLM library are provided, removing the need to create experts via prompting. Therefore, we adapt the segment of the Meta-prompting instruction dedicated to expert LLM collaboration to serve as a comparative method, which we refer to as "Meta-prompting-E" in this paper. We also compare the multi-LLM debate~\citep{du2023improving} method, as a specific case within prompting-based methods, where all the experts are invoked for all queries. In router-based methods, we compare with LLM-Blender~\citep{jiang2023llmblender}, which ranks the responses from each LLM. And we select the top-ranked result as the routing expert. In the original design of the LLM-Blender, it is also capable of fusing the top-k responses. This paper marks these two methods as LLM-Blender (Top 1) and LLM-Blender (Fused), respectively. It is noteworthy that Zooter~\citep{lu2023routing} is also a high-related router-based method. We plan to compare with it in the future since its implementation has not been released.

\paragraph{Implementations} We employ Qwen-7B-Chat~\citep{bai2023qwen} as the meta LLM. The expert token embedding is initialized with the average value of the language modeling head of word tokens for training stability. We utilize AdamW~\citep{loshchilov2019decoupled} optimizer, with a learning rate of 5e-4 and weight decay set to 1.0, for a training duration of 5 epochs. We apply greedy decoding for all comparative methods, unless specifically stated otherwise. The \ourmethod framework's training is performed on 1 $\times$ A100 GPU. The inference is conducted on 8 $\times$ A100 GPUs as it requires serving LLMs simultaneously.

\subsection{Expert Query Set Collection}

In our section, we collect the expert query set using the automatic pipeline described in \cref{subsec:learning}, on an instruction dataset generated by GPT-4 containing 15,000 query-response pairs. The instruction dataset is generated by adopting a process similar to that depicted in \cref{fig:data_collection}, using the dev set questions from all categories of MMLU as the seed question set. We also conduct a similar data cleaning process to ensure there is no data leakage in the instruction dataset. After filtering for qualifying queries, we perform down-sampling for each expert LLM, resulting in a set of 100 queries for each expert's query set.

\subsection{Evaluation of Generalist framework on MMLU-Expert}

Given a general-purpose LLM and an expert LLM library, we evaluate the performance of the generalist framework developed by comparison methods using the overall accuracy of MMLU-Expert (\cref{tab:mmlu}). From the data shown in the table, we can draw the following observations. (1) \ourmethod achieves the highest overall accuracy (73.52\%) among all methods, outperforming the runner-up method by a margin of 5.64\%.  Its performance also comes closet to the ideal scenario of selecting expert LLMs as if by an oracle. This indicates that \ourmethod is highly effective to build a generalist framework by synergizing expert LLMs across diverse domains, significantly outperforming other approaches in nearly all six domains. (2) The multi-LLM debate strategy, despite boosting overall accuracy, does not guarantee improvements across all domains. (3) The performance of the LLM-blender series methods is subpar, possibly because the ranker model designed for general scenarios is not well-suited for expert routing and collaboration. (4) The Meta-Prompting-E technique stands as the runner-up methods, which utilizes carefully crafted instructions for expert routing and collaboration, shows marked enhancements in fields like electrical engineering and international law. In \cref{subsec:expert_routing}, we delve deeper into the uneven performance across various domains. Yet, it falls short in overall accuracy compared to our approach. 

\begin{table*}[t]
\begin{center}
\caption{Accuracy (\%) on MMLU Expert dataset.}
\label{tab:mmlu}
\small
\resizebox{\textwidth}{!}{ \renewcommand{\arraystretch}{1.0}
\begin{tabular}{l|cccccc|c}
\toprule
\thead{\textbf{Method} (7 LLMs) \\ (1 Meta + 6 Experts)} & {\footnotesize Astronomy} & {\footnotesize \thead{Electrical \\ Engineering}} & {\footnotesize \thead{Human \\ Sexuality}} & {\footnotesize \thead{International \\ Law}} & {\footnotesize Prehistory} & {\footnotesize \thead{Security \\ Studies}} & \textbf{Overall}      \\ \midrule\midrule
meta LLM & 59.87 & 49.66 & 70.99 & 63.64 & 61.11 & 60.41 & 60.73 \\
Oracle Expert & 75.66 & 69.66 & 86.26 & 85.12 & 78.40 & 77.96 & 78.44 \\
\midrule
Meta-Prompting-E  & 58.55 & 64.14 & 83.21 & 72.73 & 72.22 & 59.59 & 67.89 \\ 
Multi LLM Debate  & 55.26 & 54.48 & 71.76 & 61.98 & 63.89 & 64.49 & 62.29 \\ 
\midrule
LLM-Blender (Top1) & 68.42 & 54.48 & 74.05 &  66.94 & 67.59 & 53.88 & 63.69 \\ 
LLM-Blender (Fused) & 59.87 & 46.21 & 74.05 & 64.46 & 59.26 & 44.90 & 56.80 \\ 
\midrule
\ourmethod (Ours) & 72.36 & 63.44 & 84.73 & 76.03 & 70.06 & 77.55 & \textbf{73.52}  \\
\bottomrule 
\end{tabular}}
\end{center}
\end{table*}

\begin{table}[t]
\begin{center}
\caption{Expert Routing Accuracy (\%) on MMLU-Expert dataset.}
\label{tab:router}
\small
%\resizebox{\linewidth}{!}{ \renewcommand{\arraystretch}{1.0}
\begin{tabular}{l|c}
\toprule
\thead{\textbf{Method}} & {Expert Routing Acc.}  \\ \midrule\midrule
Random & 16.67 \\ 
Oracle Expert & 100.00  \\
\midrule
LLM-Blender (Top 1)  & 28.26  \\ 
Meta-Prompting-E  & 67.08  \\ 
\midrule
\ourmethod (Ours) & 82.11 \\
\bottomrule 
\end{tabular}
%}
\end{center}
\end{table}

\subsection{Expert Routing Accuracy} 
\label{subsec:expert_routing}

In this section, we compare the expert routing accuracy in \cref{tab:router}. For example, in the MMLU-Expert dataset, the questions in electrical engineering category has an oracle expert LLM, which is fine-tuned on the synthetic electrical engineering knowledge dataset (\cref{sec:exp_setting}). The accuracy is 100\% when all the queries are routed to its corresponding domain expert LLM, where random routing has an accuracy of 16.67\%. It can be seen that the LLM-Blender, of which the ranking model is trained based on rewards aligned with human preferences, does not perform well in accurately routing to the domain expert. Meta-Prompting-E achieves a 67.08\% expert routing accuracy on the MMLU-Expert dataset, while \ourmethod reaches 82.11\%, exceeding it by 15.03\%.

\begin{figure}[h]
    \centering
    \includegraphics[width=\linewidth]{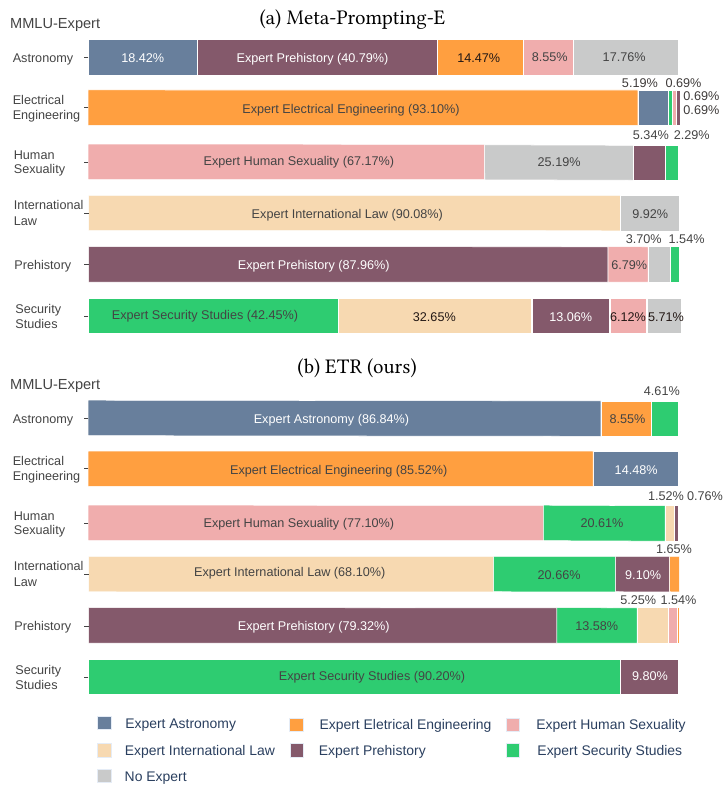}
    \caption{Routing expert distribution of Meta-Prompting-E (top) and ETR (down).}
    \label{fig:expert_dist}
\end{figure}

\cref{fig:expert_dist} visualizes the routing expert distribution of \ourmethod and Meta-Prompting-E. It can be found that the Meta-Prompting-E performs poorly in certain domains, like astronomy, where it yields a mere 18.42\% accuracy. This indicates that by merely designing crafted instructions, LLMs may struggle to accurately handle complex or implicit expertise. Comparing with the results in Table 2, it can be observed that in domains where Meta-Prompting-E's routing performance is subpar, the answer accuracy of meta-prompting is relatively low, affecting its overall performance as a generalist. In contrast, our approach maintains a routing accuracy exceeding 65\% across all six domains, demonstrating greater robustness.

\subsection{Analysis on Expert Query Set}

The construction of the expert query set is crucial for learning expert tokens with good generalizability. Ideally, the more queries in the expert query set, and the higher their quality, the more comprehensive and accurate the depiction of expertise by the learned expert tokens. However, in practical use, because the instruction dataset we use to build the expert query set automatically can be relatively small, the number of expert queries obtained may be limited. Fortunately, \ourmethod freezes the entire LLM and only tunes the expert tokens, which involves a very small number of parameters (expert number $\times$ embedding dimension). Thus, empirically we find that only 100 expert queries for each expert LLM are sufficient to achieve good results.

\begin{figure}[htbp]
    \centering
    \includegraphics[width=\linewidth]{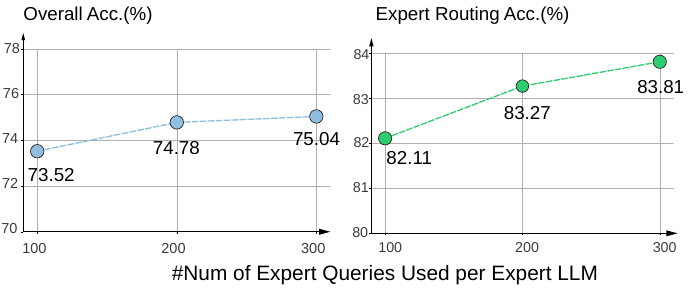}
    \caption{Overall Accuracy (Left) and Expert Routing Accuracy (Right) on MMLU-Expert as the size of the expert query set per expert increases.}
    \label{fig:query_num}
\end{figure}

\cref{fig:query_num} illustrates that when the size of the expert query set is further expanded, both the overall accuracy and expert routing accuracy of our method improve. In our comparative experiments on effectiveness, we use a expert query set size of 100 per expert LLM to simulate scenarios where expert queries are limited. However, in scenarios where expert queries are easily accessible, the performance of our method can be further enhanced.

\subsection{Dynamic Extension of Expert LLMs}

In practical applications, the expert LLM library may continuously expand with the emergence of new expert LLMs. To adapt to the extended expert LLM library, prompt-based approaches must update their instructions, whereas router-based techniques may need retraining the router model. Motivated by the weights of parameter-efficient tuning can be effectively combined~\citep{wu2023pituning}, we train the expert tokens for the newly added expert LLMs and merge newly-learned expert tokens directly into the frozen expert token head in a plug-in manner. \cref{tab:dynamic} compares the performance of the following two different settings on MMLU-Expert.

\begin{itemize}[leftmargin=2.5mm]
    \item \ourmethod-Static: All expert LLMs exist in the expert LLM library at timestep 0 (i.e., the default setting of this paper).
    \item \ourmethod-Dynamic: At timestep 0, expert tokens are trained based on the current expert LLM library. When new LLM experts are added at timestep 1, only the expert tokens of these new expert LLMs are trained and then concatenated with the previously obtained expert tokens.
\end{itemize}

Specifically, for the \ourmethod-dynamic setting, we randomly selected four expert LLMs to form the expert LLM library at timestep 0. The remaining two serve as new LLM experts to be added at timestep 1. \cref{tab:dynamic} shows that under the setting of dynamically adding expert tokens, \ourmethod suffers only a minor performance drop. This reveals an attractive feature of our framework: providers of new expert LLMs can add their trained expert LLMs and expert tokens to the framework as plug-ins, without needing to modify any other part of the framework.

\begin{table}[htbp]
\begin{center}
\caption{Performance comparison between \ourmethod-Static/Dynamic on MMLU-Expert.}
\label{tab:dynamic}
\small
\resizebox{\linewidth}{!}{ \renewcommand{\arraystretch}{1.0}
\begin{tabular}{l|c|c}
\toprule
\thead{\textbf{Variant}} & {Expert Routing Acc.} & Overall Acc.  \\ \midrule\midrule
\ourmethod-Static  & 82.11 & 73.52 \\
% \midrule
\ourmethod-Dynamic  & 81.75 (-0.26) &  73.25 (-0.27) \\
% \midrule
\bottomrule 
\end{tabular}
}
\end{center}
\end{table}

\begin{figure*}[t]
    \centering
    \includegraphics[width=\linewidth]{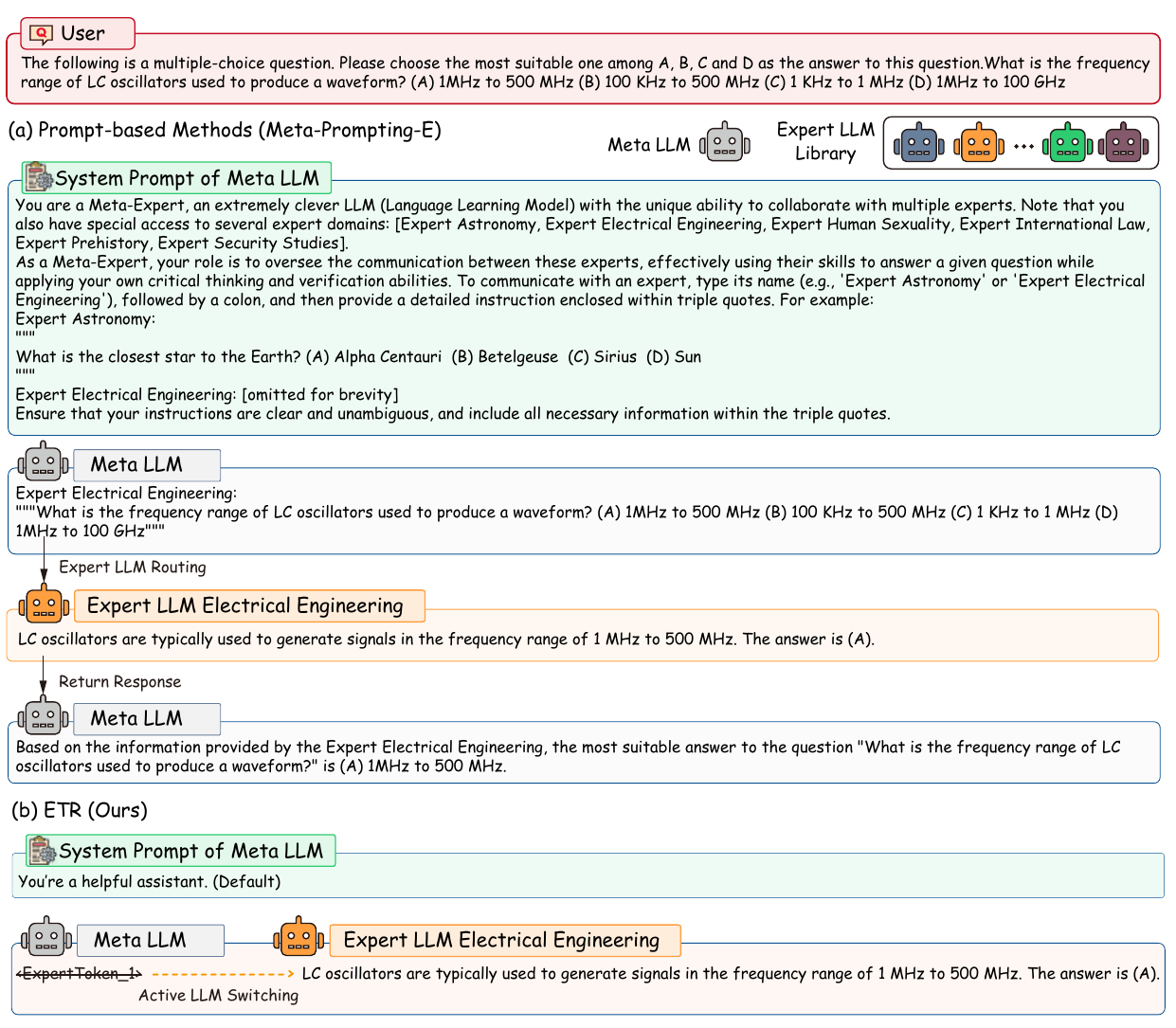}
    \caption{Comparison of running cases between prompt-based methods (Meta-Prompting-E) and \ourmethod (Ours) on MMLU-Expert dataset. <ExpertToken\_1> is the corresponding expert token to Expert LLM Electrical Engineering.}
    \label{fig:running}
\end{figure*}

\subsection{Running Case Analysis}

In \cref{fig:running}, we present a running case comparing between \ourmethod and the runner-up prompt-based method (Meta-Prompting-E). Our observations yield the following insights:

\begin{itemize}[leftmargin=2.5mm]
    \item The prompt-based method requires the design of complex instruction templates to facilitate collaboration between the meta LLM and expert LLMs. \ourmethod, however, obviates the need for using any specially designed instructions.
    \item In prompt-based methods, the information exchanging between LLMs is through conversations. While this approach is simple and effective, it adds an extra layer of complexity on the user end. In contrast, our \ourmethod framework ensures that the collaboration between LLMs remains invisible to users, allowing them to interact with our framework as if it were a single LLM.
    \item In prompt-based methods, the routing to the appropriate expert relies on the meta LLM's intrinsic capabilities in zero-shot or few-shot scenarios. In contrast, our \ourmethod framework can learn from expert queries, thus introducing new knowledge to the meta LLM, resulting in more precise expert routing.

\end{itemize}

\subsection{Running Time Analysis}

We evaluate the running time of ETR in two different conditions: no active LLM switching (line 9 in~\cref{alg:algo1}) versus active LLM switching (line 11 in Algorithm~\cref{alg:algo1}). We randomly select 100 queries to calculate the average running time. To eliminate the impact of different response length by the two conditions on the running time, the maximum generation length is set on the shorter response length of the two (greedy decoding is used for deterministic response). The experimental results are as shown in~\cref{table:time}.

\begin{table}[ht]
\centering
\caption{Time Cost of LLM Switching}
\begin{tabular}{l r}
\hline
           &  	Running Time (s)  \\ \hline
No Active LLM Switching & 1.589    \\
Active LLM Switching & 1.616    \\  \hline

\end{tabular}
\label{table:time}
\end{table}

The experiment runs with CUDA 12.1, Pytorch 2.1 and Flash-Attention 2. Each LLM is served on an A100-SXM4-80G GPU, utilizing bf16 quantization. We notice that active LLM switching results in roughly 1.7\% extra running time costs. This ratio is expected to further decrease as the generated response becomes longer.

\section{Related Works}

\paragraph{Expert LLM Collaboration} A bunch of methods aim at enhancing the reasoning capabilities of LLMs in specific tasks by instructing LLMs into experts through the crafting of prompts. These methods mainly focus on boosting a single LLM by simulating multiple expert LLM instances, which can also be applied to the setting in this paper. ExpertPrompting~\citep{xu2023expertprompting} elicits LLMs to generate expert-level responses using proper crafting of prompts. Meta-Prompting~\citep{suzgun2024metaprompting} breaks down complex tasks into subtasks handled by expert instances of a single LLM. Multi-LLM Debate~\citep{du2023improving} involves the use of multiple LLM instances to propose and debate their individual responses and reasoning processes over several rounds in order to converge on a common final answer. 

There is also a line of methods highly relevant to this paper, which involves training an additional router model to integrate multiple LLMs into one framework. LLM-Blender~\citep{jiang2023llmblender} uses a pair-ranker model for optimal LLM output selection and a gen-fuser model for merging the best outputs. Zooter~\citep{lu2023routing} is a reward-guided routing method for LLM ensembles that efficiently assigns queries to the most expert LLM. However, in scenarios where expert models are dynamically extended, these methods often require retraining the router model.

\section{Conclusion}

In this study, we introduced a novel approach \ourmethod to integrating various expert LLMs into a unified generalist framework. This method facilitates seamless collaboration among LLMs without exposing the complexity to the user, effectively turning the framework into a single, versatile LLM. Through comprehensive experiments across multiple domains, \ourmethod demonstrated superior performance over existing paradigms, offering a promising direction for enhancing LLM's generalist capabilities by combining the strengths of both general and expert models.

% \clearpage

\section{Limitations}

The main limitation addressed in this paper is the necessity for a multi-expert LLM framework to operate all expert LLMs concurrently to provide real-time services. As the number of expert LLMs grows significantly, the associated costs could become prohibitive. A potential compromise involves distilling the expert knowledge from these LLMs into more efficient, lightweight modules.

\section*{Ethics Statement}

This work has been conducted using open-source Qwen~\cite{bai2023qwen} models and aims to study methods for collaborating multiple expert language models within a singular generalist framework. We acknowledge that language models can exhibit biases and generate inappropriate content. The expert models were created using synthetic data generated by commercial language models, and data filtering was performed to avoid harmful content. Thus, we believe our work does not pose severe ethical concerns.

\section*{Acknowledgment}

This work is supported by National Natural Science Foundation of China (No. 62322606, No. 62441605) and SMP-IDATA Open Youth Fund.

\bibliography{custom}

\appendix
\clearpage

\onecolumn
\section{Appendix}
\label{sec:appendix}

\subsection{MMLU-Expert Dataset Statistic}
% \label{app:fail}
\vspace{-10pt}
\begin{table}[ht]
\centering
\caption{Dataset Statistics of MMLU-Expert}
\begin{tabular}{l r r}
\hline
Domain                   & Test Question Number & Synthetic Dataset Size \\ \hline
Astronomy                & 152   &  7,500 \\
Electrical Engineering   & 154   &  7,500 \\
Human Sexuality          & 131   &  7,500 \\
International Law        & 121   &  7,500 \\
Prehistory               & 324   &  7,500 \\
Security Studies         & 245   &  7,500 \\ \hline
\end{tabular}
\label{table:dataset_statistics}
\end{table}

\section{Hyper-parameter Setup}

\subsection{expert LLM Fine-tuning Setup}
\label{app:setup}

We provide the hyper-parameters on how different domain expert LLMs are fine-tuned in \cref{table:finetune} for reproduction. To prevent overfitting on relatively small datasets during the fine-tuning process, we employ an early stopping mechanism to save the expert LLM's checkpoints. 

\begin{table}[htb]
    \centering
    \label{table:finetune}
    \caption{Hyperparameters for expert LLM fine-tuning.}
  \resizebox{0.99\textwidth}{!}{ \renewcommand{\arraystretch}{1.0}
\begin{tabular}{lcccccc}
\toprule
Hyperparameter & Astronomy  & Electrical Engineering  & Human Sexuality & International Law  & Prehistory & Security Studies    \\
\midrule
Gradient Accumulation & 8 & 8 & 8 & 8 & 8 & 8 \\
Batch size & 16 & 16 & 16 & 16 & 16 & 16 \\
Learning Rate & $5 \mathrm{e}-5$ & $5 \mathrm{e}-5$ & $5 \mathrm{e}-5$ & $5 \mathrm{e}-5$ & $5 \mathrm{e}-5$ & $5 \mathrm{e}-5$   \\
Epochs & 5 & 8 & 5 & 8 & 5 & 6 \\
Warmup steps & 50 & 50 & 50 & 50 & 50 & 50 \\
Weight decay & $1 \mathrm{e}-1$ & $1 \mathrm{e}-1$ & $1 \mathrm{e}-1$ & $1 \mathrm{e}-1$ & $1 \mathrm{e}-1$ & $1 \mathrm{e}-1$  \\
\midrule
\ Tunable parameters & Full & Full & Full & Full & Full  & Full  \\
\bottomrule
\end{tabular}}
\end{table}

\subsection{Expert Token Training Setup}

We provide the hyper-parameters on expert token training in \cref{table:expert} for reproduction. For the sake of training stability, the initialization of the expert token is conducted using the mean value of embeddings found within the language modeling head of the meta LLM. It can be seen that when using Qwen-7B-Chat as the meta LLM, the parameters that need to be trained amount to only 24,567, which accounts for merely $3.51\times10^{-6}$ of the total parameters.

\begin{table}[ht]
\centering
\caption{Configuration of training expert tokens}
\label{table:expert}
\begin{tabular}{ll}
\hline
Hyper Parameter                     & Value                                \\ \hline
Meta Model                              & Qwen-7B-Chat                                        \\
Train Batch Size            & 16                                                              \\
Model Max Length                       & 1024                                                 \\
Learning Rate                          & 0.0005                                               \\
Weight Decay                           & 1.0                                                  \\
Number of Training Epochs              & 5                                                    \\
Learning Rate Scheduler Type           & Cosine                                               \\
Number of Warmup Steps                 & 5    \\
Expert Query Set Size                  & 600    \\ 
\# of Trainable Params.  &   24,567  \\
\hline
\end{tabular}
\label{table:simplified_training_settings}
\end{table}

\section{A Running Case of Extracting Domain Knowledge from GPT-4}

\begin{figure}[htbp]
    \centering
    \includegraphics[width=\linewidth]{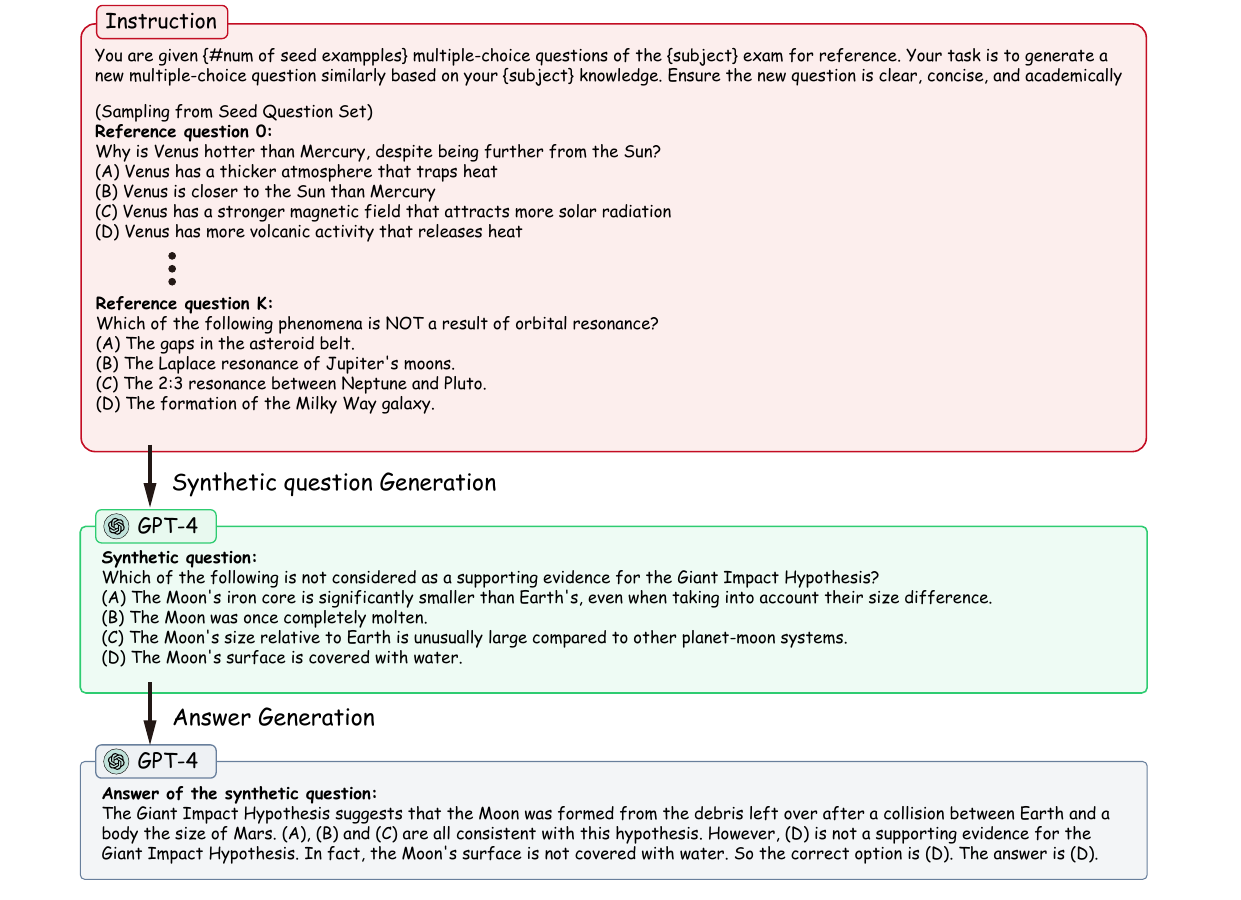}
    \caption{A running case of extracting knowledge from GPT-4 by generating synthetic questions and answers.}
    \label{fig:gpt4}
\end{figure}

In the evaluation phase, the domain-specific dataset for creating the expert model and the instruction dataset from which expert queries are extracted are both constructed using GPT-4. Thus, \cref{fig:gpt4} illustrates a example to generate synthetic questions for the extraction of domain knowledge. The generated questions are expanded into the seed question set. 

\end{document}